\def\year{2018}\relax
\documentclass[letterpaper]{article} 
\usepackage{aaai18}  
\usepackage{times}  
\usepackage{helvet}  
\usepackage{courier}  
\usepackage{url}  
\usepackage{graphicx}  
\expandafter\def\expandafter\UrlBreaks\expandafter{\UrlBreaks
  \do\a\do\b\do\c\do\d\do\e\do\f\do\g\do\h\do\i\do\j%
  \do\k\do\l\do\m\do\n\do\o\do\p\do\q\do\r\do\s\do\t%
  \do\u\do\v\do\w\do\x\do\y\do\z\do\A\do\B\do\C\do\D%
  \do\E\do\F\do\G\do\H\do\I\do\J\do\K\do\L\do\M\do\N%
  \do\O\do\P\do\Q\do\R\do\S\do\T\do\U\do\V\do\W\do\X%
  \do\Y\do\Z}
\usepackage{floatrow}
\usepackage{multirow}
\usepackage[inline]{enumitem}
\DeclareFloatFont{small}{\small}
\newcommand\blfootnote[1]{%
  \begingroup
  \renewcommand\thefootnote{}\footnote{#1}%
  \addtocounter{footnote}{-1}%
  \endgroup
}

\begin{document}
%
 \title{An Exploration of Unreliable News Classification in Brazil and The U.S.}

\author{ 
Maur\'{i}cio Gruppi, Benjamin D. Horne and Sibel Adal{\i} \\
Rensselaer Polytechnic Institute, Troy, New York, USA\\
\{gouvem, horneb, adalis\}@rpi.edu\\
}

\maketitle
\begin{abstract}

The propagation of unreliable information is on the rise in many places around the world. 
This expansion is facilitated by the rapid spread of information and anonymity granted by the Internet. 
The spread of unreliable information is a well-studied issue and it is associated with negative social impacts. 
In a previous work, we have identified significant differences in the structure of news articles 
from reliable and unreliable sources in the US media. 
Our goal in this work was to explore such differences 
in the Brazilian media. 
We found significant features in two data sets: 
one with Brazilian news in Portuguese and 
another one with US news in English. 
Our results show that features related to the writing style were prominent in both data sets and, 
despite the language difference, some features have a universal behavior, 
being significant to both US and Brazilian news articles. 
Finally, we combined both data sets and used the universal features to build a 
machine learning classifier to predict the source type of a news article as reliable or unreliable.\blfootnote{Presented at NECO 2018, co-located with ICWSM 2018}
\end{abstract}

\section{Introduction}
There is an increasing interest in developing automated tools to
identify misinformation. In our past
work~\cite{horne2017just};~\cite{horne2018accessing}, we have shown
that it is possible to distinguish between information coming from
reliable sources and unreliable sources, i.e. sources that have
published completely fabricated information.  We
have also shown that the writing style of satire and unreliable
sources have many similarities~\cite{horne2017just}, which is another
important class of articles to study given the use of humor and irony
in many extremist communities~\cite{marwick}. This work as well as many
others of similar
nature~\cite{nakashole2014language}; \cite{potthast2017stylometric}; \cite{popat2016credibility}; \cite{guacho2018semi}
concentrate on content-based prediction and analysis, illustrating the
usefulness of content approaches to automatic news classification.

Despite this growing usefulness of content-based
methods, there is little work exploring how well these methods can
generalize across language, time, and culture.  A recent study by
European Union\footnote{\url{https://goo.gl/wCzFwp}} has pointed out the need to study misinformation
identification in multiple languages and countries to gain a deeper
understanding of commonalities as well as differences. In this paper,
we provide a first attempt in classifying news sources with a unique
study of news sources from U.S. and Brazil. We ask the following
questions:

{\bf Q1:} Can we distinguish between news from reliable, unreliable
  and satire news sources based on writing style alone, both in U.S.
  and Brazilian news sources?
  
  Our objective here is twofold. We would like to revisit our
  findings and check whether they remain valid. In the past, we were
  able to show reasonably high prediction accuracy (77\% ROC AUC) with
  fairly simple features and a simple model. However, this was from a
  fairly small data set right after U.S. Elections. At the time,
  attention on news was high, but study of misinformation was just
  starting to gain momentum. Due to the recent attention on the topic
  of misinformation in news, sources may have started using different
  tactics in presenting information. While we have shown that these
  types of content features can generalize in prediction
  tasks~\cite{horne2018accessing}, we have not explored the changes in
  feature significance and direction in newer time slices.

  The Brazilian scenario is fairly susceptible to misinformation as the 
  collected data pertains to months prior to the country's presidential elections.
  Though the two domains are somewhat different, the underlying
  motivation is the same: disseminate misinformation. We study a
  similar set of features in both languages and check the accuracy of
  prediction of these features across two different domains.

{\bf Q2:} Is there a universality to some of the features? Are there
  significant features common to both countries? 

  Our second objective is to understand to which degree the prominent
  features are universal. Given two different countries with different
  political landscapes, cultural landscapes, and language, one can
  expect significant differences in how misinformation is
  presented. However, one can also expect that the main motivation of
  the sources are similar: to engage users and present information
  that appear credible. Given the universal nature of some heuristics
  used to infer credibility of information, we might expect consistent
  similarities. On top of this, some similarities may be expected
  due to lack of editorial oversight in some of the unreliable media
  organizations. Hence, to understand these similarities, we find
  significant features in both countries and check how much they
  overlap. Then, we also look at whether the differences also point in
  the same direction for these features.

First, in response to Q2 above, we find a high number of features in
multiple categories that are significant in both
countries. Furthermore, there is strong agreement in features that
measure text complexity across all three categories: unreliable
sources use simpler language, shorter texts but longer sentences than
reliable sources. These comparisons also hold for relationships
between satire sources. We also find consistent similarities for
complexity and stylistic features between reliable and unreliable sources, and weaker
overall similarity for features involving part-of-speech. In essence,
these three categories of features have a certain degree of
universality. We also test this assumption using a prediction
experiment. We show that using our fairly simple set of features, we
can distinguish reliable and unreliable sources with 85\% accuracy in
the Brazilian dataset and 72\% in the U.S. dataset. Then we combined
U.S. and Brazilian data sets for a joint classification task. We chose
the subset of features that are significant in both data sets, a total
of 18 features chosen from this universal group. The accuracy of
classification between reliable and unreliable news was 70\% using
this small set of features, illustrating the universal nature of these
features. These early promising results open many new research
questions that we expect to investigate in detail in our future work.

\section{Related Work}
In addition to work that concentrates on classification of reliable
and unreliable news based on its content, there has been several works
examining news and news consumption across countries and
cultures. Most of these works have focused on news coverage or
consumption. Vreese et al. study news coverage of the 2004 European
parliament elections in each of the EU countries, showing a difference
in sentiment towards the EU between the new and old
members~\cite{de2006news}. An and Kwak examine what gets media
attention across 196 counties using data from Unfiltered News
(\url{unfiltered.news})~\cite{an2017gets}. They find that there are
differences across region and that media has a short attention span in
general. Similarly, An et al. show that news coverage across various
countries can not only be driven by geographical closeness, but also
by historical relationships and that similarity between news coverage
in various countries depends on time and
topic~\cite{an2017convergence}. Kwak et al. examine both the attention
of media and the attention of consumers through a unique study of news
coverage and news searches in 193 countries. They show that many
countries have dissimilar attention between media an public attention,
but local attention patterns are similar. These differences in news
coverage across countries may explain some of the differences in
content we find this this study. Despite the interest in cross country
news coverage and consumption, there is little to no work exploring
news content differences and similarities across languages or
countries, especially in the context of unreliable news. In addition,
we do not find many studies exploring the generalization of
content-based features for prediction of reliable and unreliable
content across cultures and contexts.

\section{Data and Features}
To study the problem of identification of articles from reliable (R),
unreliable (U), and satire (S) sources, we construct two sets of
political news articles from US (United States) and BR (Brazilian)
sources in each category.  Reliable sources in each country are
well-established media companies. Unreliable sources are sources known
to have published at least one maliciously incorrect news article
(according Snopes in US and AosFatos (\url{aosfatos.org}) in BR). To
this group, we also add sources that self-identify as satire and
clearly indicate this on their website. The sources in US data comes
from the sources in the NELA2017 data set~\cite{horne2018sampling}.
We construct BR sources by looking for unreliable and satire media
sources and well-established media companies. We collect all political
articles from these sources for a period of one month, between February 15th and March 15th of 2018, and then sample
articles from each source.


Our BR news dataset contains 5511 political news articles from 19
sources of which 4698 articles are from reliable sources, 755 are from
unreliable sources and 58 from satire sources.  The list of sources is
shown in Table \ref{tab:br_sources}.  Our US news dataset contains
2841 political news articles from 16 sources of which 1997 articles are
from reliable sources, 794 are from unreliable sources and 50 are from
satire sources.  The list of US news sources is shown in
Table~\ref{tab:us_sources}.  Both BR and US datasets contain all
articles collected between February 15th and March 15th of 2018 from
the aforementioned sources.  Each article is a data point in our
dataset.  For each article, we compute every feature from our feature
list, and assign a class \emph{Reliable (R)}, \emph{Unreliable (U)}
or \emph{Satire (S)} based on the source from which the article was
collected.

We construct a set of roughly equivalent sets of features in both
languages as shown in Table~\ref{tab:features}.  The features are
classified into 4 categories: complexity, style, linguistic, and
psychological. Each feature is computed on title and body text
separately.  Some of these features are obtained using the Python
NLTK \cite{bird2006nltk} and LIWC \cite{pennebaker2001linguistic}.

\textbf{Complexity features} are used to assess the level of intricacy
of title and body text of news articles. We capture the sentence level
complexity through the number of words per sentence. To capture the
readability level of the text, we use the Gunning fog index, SMOG
grade, Flesch-Kincaid grade level and Flesch-Kincaid reading ease
indexes. Such scores suggest the education level needed for the reader
to have some understanding of the text. Higher scores indicate that a
higher education is required. 

\textbf{Stylistic features} are related to the writer's style at the
character level.  These features include the frequency of commas and
 punctuation, the number of words in all caps. It is common to see use
 of capitalization and exclamation points in sensationalist writing
 styles. Journalistic style in contrast is much more measured in its
 use of these stylistic features. One can also see stylistic
 differences due to the lack of clear editorial oversight and
 standards in alternative media sites.

\textbf{Linguistic features} are related to the frequency of different
 parts of speech used in the text, such as frequency of nouns, proper
 nouns, verbs, etc. These features often indicate how the text is
 framed, such as whether the article is about specific individuals or
 actions, or it is from the point of view of a specific person. 

\textbf{Psychological features} are based on words correlated to
 psychological processes, such features are provided by the Linguistic
Inquiry and Word Count dictionaries (in English and Portuguese).
These a non-topic related features that evoke cognitive processes such
as positive and negative emotions, anxiety, certainty, etc.

\begin{table}[]
\centering
\fontsize{8}{8.5}\selectfont
\caption{Brazilian news sources}
\label{tab:br_sources}
\begin{tabular}{|c|c|c|}
\hline
\textbf{Reliable (R)}     & \textbf{Unreliable (U)}        & \textbf{Satire (S)}  \\ \hline
BBC Brasil        & Correio do Poder     & Joselitto M\"{u}ller \\
El Pa\'{i}s Brasil    & Di\'{a}rio do Brasil     & Sensacionalista  \\
Exame             & Folha Pol\'{i}tica       & Piau\'{i} Herald     \\
Extra             & Gazeta Social        &                  \\
Folha de S. Paulo & Jornal do Pa\'{i}s       &                  \\
G1                & Pensa Brasil         &                  \\
Isto \'{E}            & Sa\'{u}de Vida e Fam\'{i}lia &                  \\
O Tempo           &                      &                  \\
Reuters Brasil    &                      &                 \\
\hline
\end{tabular}
\end{table}

\begin{table}[]
\centering
\fontsize{8}{8.5}\selectfont
\caption{US news sources}
\label{tab:us_sources}
\begin{tabular}{|c|c|c|}
\hline
\textbf{Reliable (R)} & \textbf{Unreliable (U)} & \textbf{Satire (S)} \\ \hline
CBS News & Activist Post & Glossy News \\
CNBC & Addicting Info & The Borowitz Report \\
NPR & Infowars & The Burrard Street Journal\\
Reuters & Intellihub & The Spoof \\
The NY Times & Natural News &  \\
USA Today & Waking Times & 				\\ 
\hline
\end{tabular}
\end{table}

\begin{table*}[]
\centering
\fontsize{7}{8}\selectfont
\caption{Features used in this study grouped by category}
\label{tab:features}
\begin{tabular}{llllll}
\textbf{Abbr.}       & \textbf{Description}                 & \textbf{Abbr.} & \textbf{Description}   &      \textbf{Abbr.} & \textbf{Description}             \\ \hline
\multicolumn{2}{c}{\textbf{Category 1: Complexity features}}  & \multicolumn{2}{c}{\textbf{Category 3: Parts of speech features}}    & \multicolumn{2}{c}{\textbf{Category 4: Psychological features}}\\
GI                   & Gunning fog grade readability index & Pronoun              & Frequency of pronouns                & Insight        & Frequency of insight related words          \\
SMOG                 & SMOG readability index & PPronoun             & Frequency of proper pronouns         & Percept        & Frequency of perceptual process words       \\
FK-RE                & Flesch-Kincaid reading ease index & IPron                    & Frequency of I pronoun               & Posemo         & Frequency of positive emotion words         \\
FK-GL                & Flesch-Kincaid grade level & You                  & Frequency of you pronoun             & Tentat         & Frequency of tentative words                \\
TTR                  & Type-Token Ratio (lexical diversity)  & SheHe                & Frequency of pronouns she and he     & Negemo         & Frequency of negative emotion words         \\
WC                   & Word count & We                   & Frequency of pronoun we              & Certain        & Frequency of certainty words                \\
WPS                  & Words per sentence   & Negate               & Frequency of negation words          & Sad            & Frequency of words related to sadness       \\ 
AVG\_WLEN            & Avg. length of words & Compare              & Frequency of comparison words        & Achieve        & Frequency of achievement words             \\  
SixLtr               & Frequency of six letter words & Preps                & Frequency of prepositions            & Anger          & Frequency of anger words \\
\multicolumn{2}{c}{\textbf{Category 2: Stylistic features}} & Article              & Frequency of articles                 & AllPunc        & Frequency of punctuation characters          \\  
Comma          & Frequency of commas & Verb                 & Frequency of verbs                    & Anx            & Frequency of anxiety words \\
 Exclam         & Frequency of exclamation marks  & AuxVerb               & Frequency of auxiliary verbs                & Cause          & Frequency of causal words (because, effect) \\
Quote          & Frequency of quotations & Quant                & Frequency of quantifying words        & Discrep        & Frequency of discrepancy words \\
Period         & Frequency of period characters & Number               & Frequency of numerals               & Feel           & Frequency of feeling words                  \\   
QMark          & Frequency of question marks & Adjective            & Frequency of adjectives         & \\     
Parenth        & Frequency of parentheses & Conj                 & Frequency of conjunctions       & \\ 
AllCaps        & Frequency of words in all capital letters & & \\
\hline
\end{tabular}
\end{table*}

\begin{table*}[h]
\fontsize{8}{8.5}\selectfont
\centering
\caption{Agreement between datasets BR and US (the agreement with respect to Kendall-tau is in the range (1,-1). TTL is title, TXT is body text.}.
\label{tab:category1agree}
\hspace{-0.5cm}\begin{tabular}{ccc}
\begin{tabular}{|l|l|l|l|} \hline
\textbf{Feature} & \textbf{Where} & \textbf{BR}                                & \textbf{US}                                \\ \hline
SMOG             & TXT       & U \textgreater R \textgreater S & U \textgreater R \textgreater S \\
GF               & TXT       & U \textgreater R \textgreater S & U \textgreater R \textgreater S \\
FK-RE            & TXT       & S \textgreater R \textgreater U & S \textgreater R \textgreater U \\
FK-GL            & TXT       & U \textgreater R \textgreater S & U \textgreater R \textgreater S \\
WC               & TXT       & R \textgreater U \textgreater S & R \textgreater U \textgreater S \\
WPS              & TXT       & U \textgreater R \textgreater S & U \textgreater R \textgreater S \\ \hline
FK-RE            & TTL       & U \textgreater S = R            & S \textgreater R \textgreater U \\
WC               & TTL       & U \textgreater S \textgreater R & U \textgreater R \textgreater S \\
WPS              & TTL       & U = S \textgreater R            & U \textgreater R \textgreater S \\
TTR              & TTL       & S = R \textgreater U            & R \textgreater S = U           \\ \hline 
\end{tabular}
&
\begin{tabular}{|l|l|l|l|} \hline
\textbf{Feature} & \textbf{Where} & \textbf{BR}                                & \textbf{US}                                \\ \hline
AllCaps          & TXT            & R = U = S                       & S \textgreater R \textgreater U \\
Colon            & TXT            & U \textgreater S = R            & U \textgreater S = R            \\
QMark            & TXT            & S \textgreater U \textgreater R & S \textgreater U \textgreater R \\
Exclam           & TXT            & U = S \textgreater R            & S \textgreater U \textgreater R \\
Dash             & TXT            & R = U \textgreater S            & S = R \textgreater U            \\
Parenth          & TXT            & R \textgreater U \textgreater S & U \textgreater S \textgreater R \\
OtherP           & TXT            & U \textgreater R \textgreater S & U \textgreater R \textgreater S \\ \hline
AllCaps          & TTL            & U \textgreater R \textgreater S & S = U \textgreater R            \\
SixLtr           & TTL            & S = R \textgreater U            & U = R \textgreater S            \\
Colon            & TTL            & U \textgreater R \textgreater S & U \textgreater R = S            \\
SemiC            & TTL            & U = R \textgreater S            & R \textgreater S = U            \\
Exclam           & TTL            & U \textgreater R = S            & S \textgreater U \textgreater R \\ \hline
\end{tabular}
&
\begin{tabular}{|l|l|l|l|} \hline
\textbf{Feature} & \textbf{Where} & \textbf{BR}                                & \textbf{US}                                \\ \hline
Funct            & TXT            & S \textgreater U \textgreater R & S \textgreater R \textgreater U \\
Pronoun          & TXT            & S \textgreater U \textgreater R & S \textgreater R \textgreater U \\
PPronoun         & TXT            & S \textgreater U \textgreater R & S \textgreater R \textgreater U \\
SheHe            & TXT            & S \textgreater U \textgreater R & S \textgreater R \textgreater U \\
IPron            & TXT            & S \textgreater U \textgreater R & S \textgreater R \textgreater U \\
Article          & TXT            & S \textgreater U \textgreater R & S \textgreater R \textgreater U \\
AuxVerb          & TXT            & S \textgreater U \textgreater R & R \textgreater U = S            \\
Negate           & TXT            & U = S \textgreater R            & U \textgreater R \textgreater S \\
Quant            & TXT            & S \textgreater U \textgreater R & U = S = R           		\\ \hline           
\end{tabular} \\
(a) Category 1 & (b) Category 2 & (c) Category 3 \\
Overall agreement: 0.5 & Overall agreement: -0.03 & Overall agreement: 0.13 \\
Unreliable vs Reliable agreement: 0.9 & Unreliable vs Reliable agreement: 0.58 & Unreliable vs Reliable agreement: 0.11 \\
\end{tabular}
\end{table*}

\section{Methodology}
To reduce the dimensionality and find the most significant features,
we performed hypothesis testing using the one-way ANOVA (ensuring our
feature distribution are normal).  First, for each dataset (BR and
US), we separate our data into three classes: reliable, unreliable and
satire news articles.  Then, for each feature, we apply a hypothesis
test to the distributions of that feature over each pair of classes
(reliable vs.~unreliable; reliable vs.~satire; unreliable vs.~satire).
If the tests result in a low p-value ($< 0.05$), the distributions of
the tested feature over different classes is statistically
significant.  To measure the effect of that significance, we use
Cohen's d effect size. The effect size quantifies the difference
between the two distributions, a large effect size implies the values
of the feature are considerably different across the classes. If a
feature has a small p-value and a large effect size, we say the
feature is \emph{significant}.

We select the most significant features in the dataset to run a
Support Vector Machine (SVM) classifier with a linear kernel. Features are selected according to Cohen's d
magnitude descriptor, features whose effect size magnitude is at least
$0.5$ (medium) are included used in the SVM. The unbalanced number of
samples in each class is handled by upsampling the data in the least
populated class, this gives us a baseline accuracy of $50\%$. We use
three binary classifiers to separate Reliable vs Unreliable articles.

To assess the universality of features, we compared the ordering
relations between distributions of features in the three classes
between US and BR using Kendall-tau where for each pairs of classes
from R,U and S, we count agreements (+1) and disagreements (-1) in the
orderings, add the numbers and divide by total number of
comparisons. Kendall-tau ranges between +1 (complete agreement) and -1
(complete disagreement). To obtain rankings between classes R, U and S
for each feature, we first check the effect size. If its magnitude is
below the $0.2$ threshold, the classes are expected to have similar
values for that feature (i.e. equality). Otherwise, we order classes
based on their expected value.

\section{Results}
We show the ordering of significant features for Categories 1, 2 and 3
in Table~\ref{tab:category1agree} (a), (b) and (c).  First, we observe
significant similarities in Category 1 for all U, R and S. Unreliable
articles use simpler language, are shorter overall but have longer
sentences than reliable articles. Unreliable article titles are longer
than reliable article titles, this suggests that unreliable sources try to convey 
as much information as possible in the title, in an attempt to draw the reader's attention. 
Satire in contrast uses more complex
language but is shorter in general than all groups. It is clear
unreliable articles use simpler language to be easily understood and
try to convey their message through longer sentences. In the stylistic
features of Category 2, the agreement is low overall, but reasonably
high with respect the comparison between reliable and unreliable
sources. Unreliable sources use more question marks, exclamation
points, all caps both in body and title, revealing the usage of more informal language. 
Ultimately, we believe the main purpose of these is to get attention of readers. 
Category 3 shows good overall agreement, but lower agreement regarding reliable and
unreliable sources. The main source of agreement in this case is with
the satire articles compared to other articles. In essence, the parts
of text features are more useful for categorizing satire, but not
unreliable articles. Such agreements tell us which features are relevant for both datasets simultaneously, 
thus displaying universality among the languages. These are the features that can be used by our 
machine learning model to classify sources in a dataset that contains both BR and US news articles. 
Furthermore, these results indicate that, in both datasets, stylistic and complexity features play an important 
role in separating articles according to source type.

The classification of BR and US datasets used the 60 and 49 most relevant features,
respectively. The test accuracy for the BR and US datasets were 85\%
and 72\%, respectively, with a baseline score of 50\%. The combined
dataset used a reduced set of features consisting of an intersection
of the most relevant features observed in both BR and US, achieving a
test score of 70\% using only 18 features. See Table \ref{tab:classification}. The majority of significant
features in the combined dataset were from categories 1 and 2
(complexity and stylistic). This result reinforces our previous
findings which show how writing style is substantially different
between reliable and unreliable news sources. Furthermore, the results
also suggest the existence of universality of complexity and stylistic
features when separating reliable from unreliable articles in both
Portuguese and English languages.


\begin{table}[]
\centering
\caption{Classification test scores for classifying R vs U in the BR, US, and combined BR + US dataset. The baseline score is 50\%.}
\label{tab:classification}
\begin{tabular}{|c|c|c|} \hline
\textbf{BR} & \textbf{US} & \textbf{BR + US} \\ \hline
85\%        & 72\%        & 70\% 			\\ \hline            
\end{tabular}
\end{table}

\section{Conclusion and Future Work}
In this study, we strengthen the claim that suggests the existence of noticeable differences between news articles from reliable and unreliable sources. Writing style and complexity are extremely significant for distinguishing between articles of the two classes. We have shown that these features may be used to classify news articles in a language other than English. In addition, we have found evidence of universality of such features across the Portuguese and English languages by using a single set of features in the classification of a combined dataset containing articles from BR and US datasets and achieving fair classification accuracy. In our future work, we intend to expand the exploration to other languages that may share commonalities in the separation of reliable and unreliable sources and carry out this experiment on different time frames, such as in mid and post-election periods, to evaluate the effects of temporal dynamics over the study. We hope these results contribute to develop guidelines for identification of sources of unreliable information.


\bibliographystyle{aaai}
\bibliography{references}

\end{document}